\documentclass[letterpaper]{article}

\usepackage{natbib,alifeconf}  
\usepackage{booktabs}
\usepackage{float}
\usepackage{subcaption}
\usepackage{listings}
\usepackage{tcolorbox}
\usepackage{amsmath}
\usepackage{url,hyperref,cleveref}

\usepackage{array}

\title{Problem-Solving in Language Model Networks}

\author{Ciaran Regan$^{1}$, Alexandre Gournail$^{2}$, Mizuki Oka$^{1}$\\
\mbox{}\\
$^1$University of Tsukuba, Japan\\
$^2$Grenoble INP - Ensimag, UGA, France\\
mizuki@cs.tsukuba.ac.jp
} 

\begin{document}

\maketitle

\begin{abstract}
    To improve the reasoning and question-answering capabilities of Large Language Models (LLMs), several multi-agent approaches have been introduced. While these methods enhance performance, the application of collective intelligence-based approaches to complex network structures and the dynamics of agent interactions remain underexplored. This work extends the concept of multi-agent debate to more general network topologies, measuring the question-answering accuracy, influence, consensus, and the effects of bias on the collective. The results show that random networks perform similarly to fully connected networks despite using significantly fewer tokens. Furthermore, a strong consensus among agents correlates with correct answers, whereas divided responses typically indicate incorrect answers. Analysing the influence of the agents reveals a balance between self-reflection and interconnectedness; self-reflection aids when local interactions are incorrect, and local interactions aid when the agent itself is incorrect. Additionally, bias plays a strong role in system performance with correctly biased hub nodes boosting performance. These insights suggest that using random networks or scale-free networks with knowledgeable agents placed in central positions can enhance the overall question-answering performance of multi-agent systems.
\end{abstract}

\begin{figure*}[t!]
    \centering
    \includegraphics[width=0.9\linewidth]{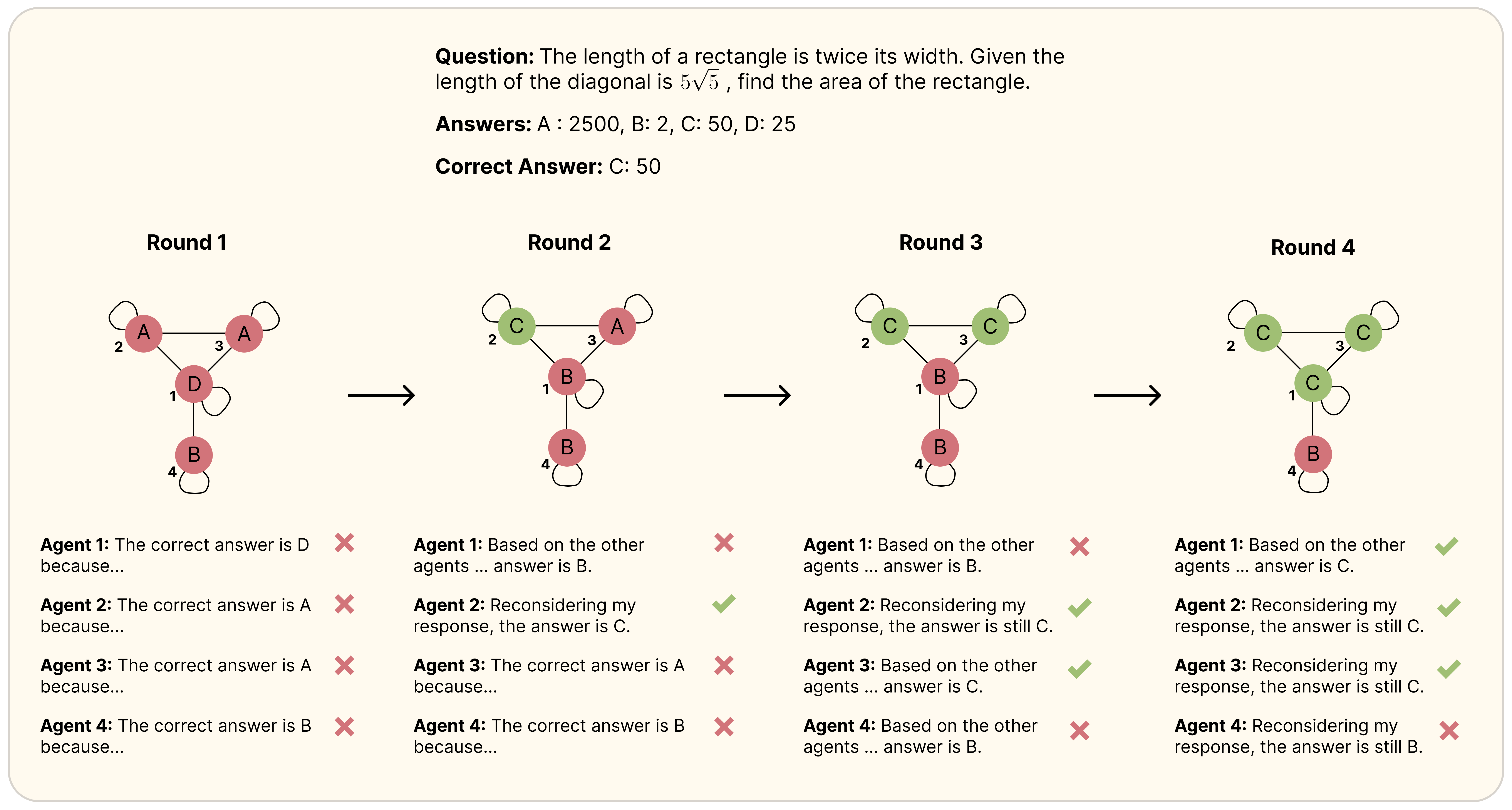}
    \caption{An overview of multi-agent debate on networks. Each node represents an agent and each edge represents a communication channel between agents, with self-loops indicating agent self-reflection. In the first round, each agent answers the question individually, with all the agents getting the answer incorrect. In the second round, agent 2 gets the answer correct through self-reflection. This correct answer then spreads through the network in subsequent rounds of debate. After the last round of debate, ``C'' is taken as the final answer of the system as this is the most common answer.}
    \label{fig:architecture}
\end{figure*}

\section{Introduction}
Large Language Models (LLMs) have demonstrated impressive performance on a wide range of tasks, such as reasoning and question-answering (QA), however, they are still prone to hallucinations and fallacious answers. To address these issues, a wide range of techniques have been introduced, many of which have been motivated by the ways in which humans approach problem-solving. Inspired by the ability of humans to combine task-oriented actions with verbal reasoning, ReAct \citep{react} enables LLM instances, or agents, to engage in a series of reasoning and action steps, which leads to a reduction in hallucinations in QA tasks. Similarly, motivated by how humans iteratively learn from past mistakes, Reflexion~\citep{reflexion} introduces agent self-reflection, where agents reflect on their past responses to induce better decision-making in future trials. 

While these approaches improve performance, they make use of only a single agent. Collective intelligence, on the other hand, suggests that multi-agent approaches can be used to create systems which are more adaptable and robust~\citep{ha2022collective}. To this end, several multi-agent methods have been proposed, such as ~\citep{moreAgents}, which suggests that scaling the number of agents increases the overall performance in problem-solving tasks. In this approach, many agents individually solve problems, before a majority vote is carried out to determine the final answer of the system. Although this technique introduces multiple agents, the agents do not interact with each other. In contrast, multi-agent debate ~\citep{debate, liang2023encouraging} combines the idea of collaborative problem-solving with agent self-reflection. In this approach, agents first solve problems as individuals, before they answer the question again after considering both their previous response and the responses of all other agents. This process repeats for several rounds before a majority vote is taken to determine the final answer of the system. This method has been shown to increase the QA performance when compared to single-agent baselines, with the accuracy increasing with the number of rounds and the number of agents.

Although ~\citep{debate, moreAgents} demonstrates that agent collaboration is a useful approach to improve QA performance, it is not clear how the topological structure of the system impacts this performance. Furthermore, the dynamics of how these agents influence each other are yet to be understood. To this end, this work generalises multi-agent approaches to more complex network structures by describing the system as an undirected graph, where each agent is represented by a node, connected to their debate partners along communication channel edges. In this context,~\citep{debate} and ~\citep{moreAgents} can be considered fully connected and fully disconnected networks respectively, with self-loops indicating agent self-reflection. Indeed, it is not clear how this approach extends to other types of networks. While networks of LLM agents have been studied in the context of opinion dynamics in social systems ~\citep{gao2023s, chuang2023simulating}, these complex network structures are yet to be applied to the problem-solving domain.

In this work, multi-agent problem-solving is implemented on scale-free, random, fully connected and fully disconnected networks, with the latter two serving as a benchmark to ~\citep{debate,moreAgents}. Studying complex networks in this context is motivated by one of the main limitations of agent debate on fully connected networks; as the number of agents increases so too does the input prompt for each agent, as the responses of all agents are aggregated together. Not only do agents begin to lose focus on the entire input prompt~\citep{debate}, but this approach becomes computationally expensive in large systems due to the growing number of connections. In a fully connected network of $n$ agents, the number of edges is given by $\frac{n(n-1)}{2}$, resulting in a quadratic increase in the number of input tokens as the number of agents increases. Additionally, the limited context window of current LLMs makes this approach infeasible with a large number of agents, highlighting the need for alternative network topologies with different degree distributions. Scale-free networks are of particular interest in this study, as the presence of hubs provides an opportunity to investigate the importance of central, well-connected nodes in the problem-solving domain. Furthermore, although there is some debate about the real-world occurrence of scale-free networks ~\citep{barabasi1999emergence, broido2019scale}, studying multi-agent problem-solving within these structures may provide valuable insights into the design of social systems.

To understand how network topology affects the QA performance, we first measure the QA accuracy by administering mathematics questions to various 25-agent networks. We then explore the effect of bias in these networks by manually inserting correct or incorrect answers into hubs or edges of scale-free networks, providing insight into how biased nodes impact the collective. The agent's influence is then examined, by analysing how likely an agent is to agree with their neighbours, and how likely they are to change their responses from their previous answer, thus providing insight into the relative strength of individuality versus collaborative problem-solving. Additionally, the dynamics of the agent's responses are investigated, to understand if agents tend to remain correct or switch back and forth between answers. Finally, we analyse the level of consensus in the networks, to understand how split the responses are within the system.

Our results show that random networks achieve similar performance to fully connected networks while using significantly fewer input tokens. In contrast, fully disconnected networks, characterised by agent self-reflection only, demonstrate inferior performance, with agents' answers degrading throughout the debate. Additionally, the tendency for connected networks to reach a strong consensus when the answer is correct, but disagree otherwise, provides a way to quantify uncertainty in the system. Furthermore, analysing biased systems suggests that agents are easily influenced by stubborn, well-connected agents.

Understanding the dynamics of these systems is an important topic in the field of artificial life ~\citep{bedau2000open, bullock2023agent,valentini2018transfer, khaluf2019modulating}, and analysing language model networks has significant implications for how we design future models of collective intelligence. In particular, we believe that utilising random networks provides a cost-effective way to further improve the problem-solving capabilities of LLMs, with the level of consensus acting as a measure of uncertainty. Moreover, the increase in performance when correct agents are at the hubs of scale-free networks suggests that future systems may benefit from having powerful models at their core, and smaller models at their periphery.

In summary, this work makes the following contributions:

\begin{itemize}
    \item We extend the concept of multi-agent debate to more complex network topologies, demonstrating that random networks perform similarly to fully connected networks while using significantly fewer tokens.
    \item We show that biased agents significantly affect the overall QA performance, especially when correct agents are positioned at network hubs, highlighting their impact on the collective.
    \item We show the importance of both individuality and collaboration, illustrating how self-reflection and interactions with neighbours influence agent behaviour. 
    \item We demonstrate that agents tend to agree when the system answers correctly but are divided otherwise, quantifying the system's uncertainty.
\end{itemize}

\section{Methods}

By representing LLM agents as nodes which are connected to their debate partners along communication channel edges, multi-agent debate can be extended to complex network topologies. Formally, a multi-agent system can be represented as a graph $G = (A,E)$ where $A = {a_1, a_2, ..., a_n}$ is a set of $n$ debating agents and $E \subseteq \{(a_i, a_j) | a_i, a_j \in A  \}$ is the set of communication links between debate partners. Notably, the existence of self-loops in these graphs allows for self-reflection.

\begin{figure}[!ht]
    \centering
    \begin{subfigure}{\linewidth}
        \includegraphics[width=1\linewidth]{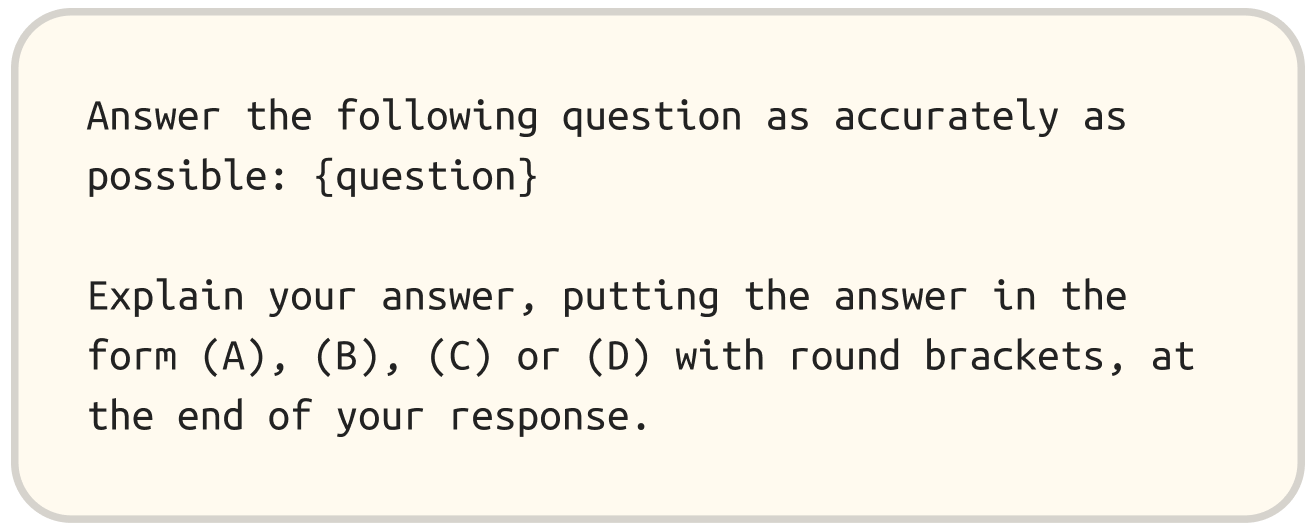}
        \caption{Initial QA prompt. The agent is instructed to answer a multiple-choice question and explain their reasoning. Here $\texttt{question}$ refers to a multiple choice question with answers A, B, C or D.}
        \label{fig:initial_prompt}
    \end{subfigure}
    
    \medskip 
    
    \begin{subfigure}{\linewidth}
        \includegraphics[width=1\linewidth]{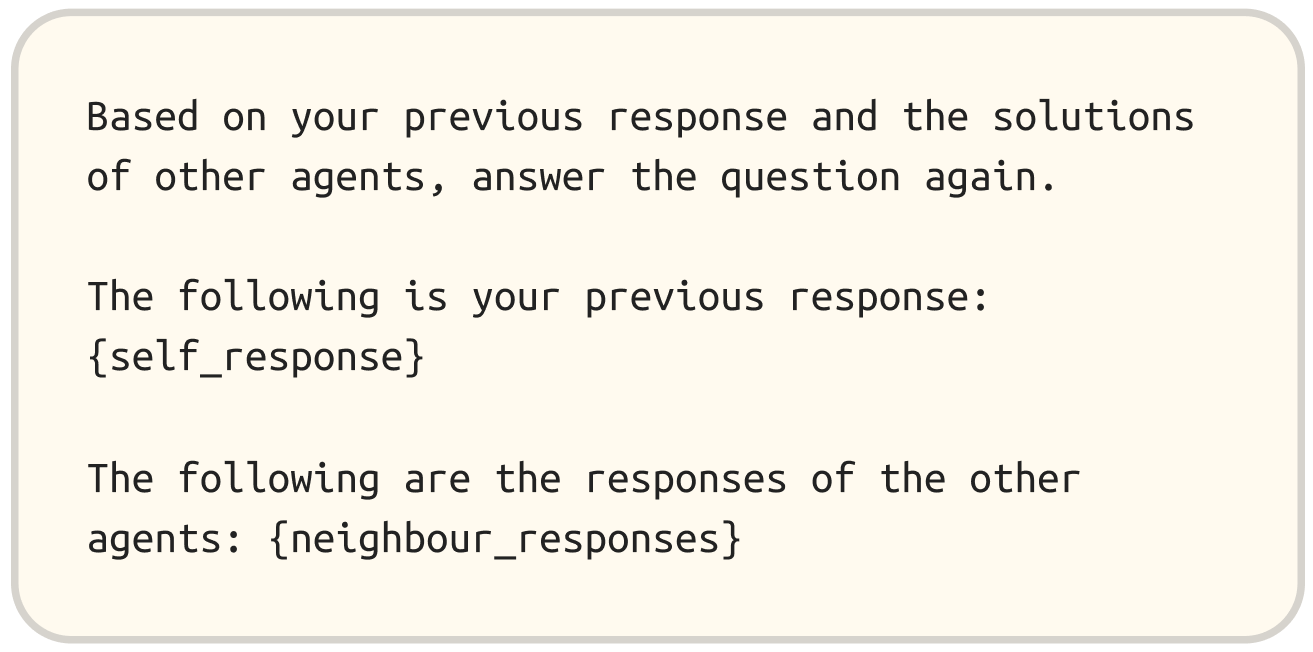}
        \caption{Follow-up prompt for the agent to reconsider their answer after being exposed to their own and others' responses. This prompt is concatenated with (a) for QA in follow-up rounds. Here $\texttt{self\_response}$ refers to the previous response of the individual and $\texttt{neighbour\_responses}$ refers to the concatenation of the neighbours' responses.}
        \label{fig:follow_up_prompt}
    \end{subfigure}
    
    \caption{The two-part prompt used for question answering. Subfigure (a) presents the initial prompt for agents to solve the problem independently. Subfigure (b) introduces the second stage, where agents are asked to re-evaluate their response after considering peer feedback and their previous response.}
    \label{fig:question_prompt}
\end{figure}

In this approach, agents first solve the problem individually. Then, each agent re-evaluates their solution by considering the responses and reasoning provided by neighbouring agents, along with their own previous solution. Importantly, the reasoning behind these solutions is included in the subsequent iteration, enabling agents to incorporate both their own and others' thought processes when revising their solutions. This process repeats for several rounds, until a majority vote is carried out, and the most popular answer is taken as the answer of the collective. By querying the system with a wide range of questions, the overall QA performance can be estimated by calculating the percentage of questions the system solved correctly. An overview of this approach is shown Fig~\ref{fig:architecture}, with the prompt used for implementing QA shown in Fig~\ref{fig:question_prompt}.

\begin{figure}[]
\includegraphics[width=1\linewidth]{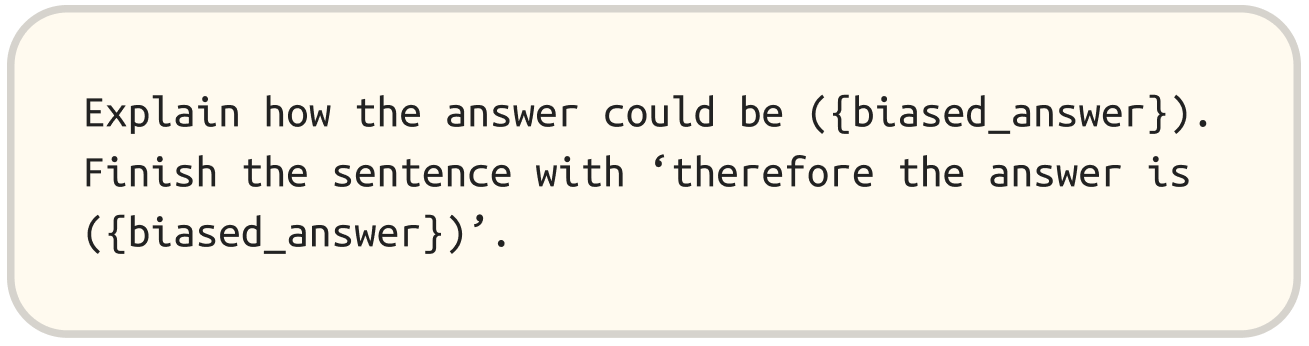}
\caption{The prompt used to generate reasoning for biased agents, where $\texttt{biased\_answer}$ is either the correct answer for correctly biased agents or an incorrect answer for incorrectly biased agents.}
\label{fig:bias_prompt}
\end{figure}

In addition to evaluating the QA performance across different network topologies, this study also investigates how the system responds to the introduction of bias; where several agents are specifically instructed to answer either correctly or incorrectly. By placing biased agents at the hubs or edges of scale-free networks, the effect on the performance is analysed, providing insights into how agents influence their neighbours. To administer this bias, agents are specifically provided with the correct or incorrect answer and asked to create a potential justification for this answer, with the prompt shown in Fig~\ref{fig:bias_prompt}. Crucially, this prompt generates an explanation for this biased answer, even if it is incorrect, which allows for fair comparison in this debate system where agents share their reasoning. Throughout the debate, the biased agent's response will not change from this initial answer. 

\section{Experimental Setup}
\begin{figure}[]
    \centering
    \begin{subfigure}[b]{0.8\linewidth}
        \centering
        \includegraphics[width=\linewidth]{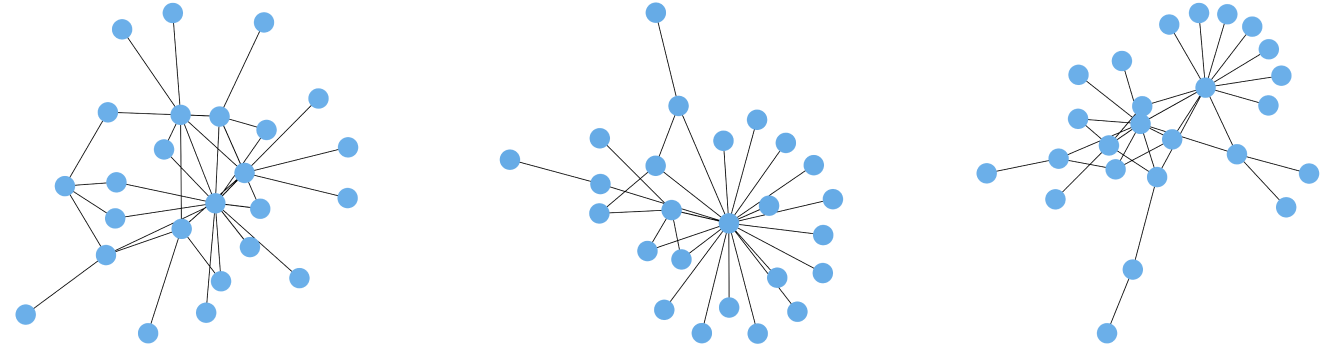}
        \caption{Scale-Free Networks}
        \label{fig:network_a}
    \end{subfigure}
    
    \begin{subfigure}[b]{0.8\linewidth}
        \centering
        \includegraphics[width=\linewidth]{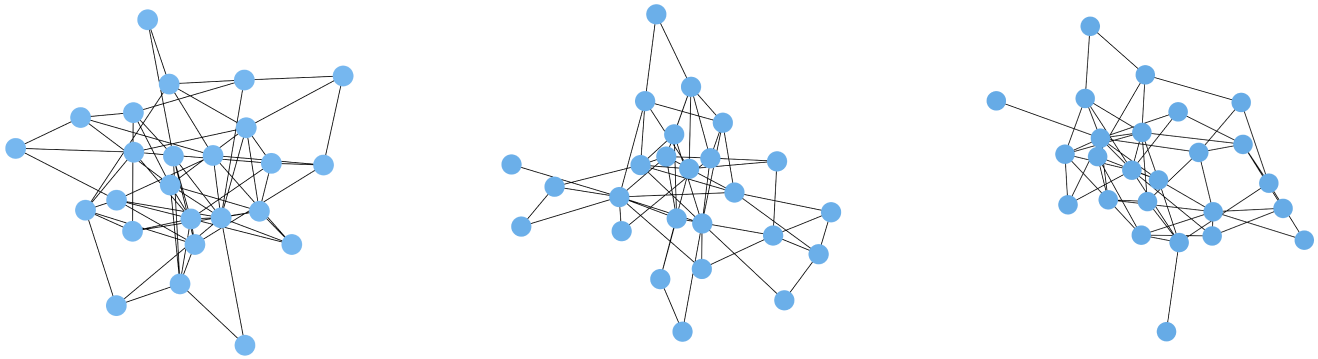}
        \caption{Random Networks}
        \label{fig:network_b}
    \end{subfigure}
    
    \caption{The specific scale-free and random networks used in the experiments.}
    \label{fig:networks}
\end{figure}

To analyze the QA performance and dynamics of these systems, 3 scale-free and 3 random 25-agent networks were used, shown in Fig~\ref{fig:networks}, in addition to fully connected and fully disconnected networks. In particular, the scale-free 
and random networks were generated using the algorithms proposed by ~\citep{bollobas2003directed} and ~\citep{gilbert1959random}, respectively, which have been implemented in the NetworkX Python library ~\citep{hagberg2008exploring}. The agents, powered by GPT-3.5-Turbo, then engaged in 4 rounds of debate, answering 100 questions from the MMLU high school mathematics dataset ~\citep{hendryckstest2021}. To keep the reasoning and answers concise, the agents were limited to output a maximum of 200 tokens each. To measure the QA accuracy of the collective, we take the most common answer at the end of the debate to be the response of the system. The average number of times the system answered a question correctly can then be calculated to give an estimate of the performance. Moreover, each of the 100 questions were administered 3 times, enabling the average accuracy of the system to be measured to a sufficient certainty. As the average diameter of the network used is approximately 4, 4 rounds of QA should be enough for sufficient information spread. These choices of parameters are further discussed in the Discussion and Limitations section.

To understand how biased hub nodes affect performance, the two most central nodes, measured by degree centrality, in each scale-free network were biased with either the correct or an incorrect answer. A similar biasing was applied to two random edge nodes for comparison. These agents do not change their response throughout the debate, and therefore provide a consistently biased answer to each of their neighbours. The debates are then run as before, allowing the performance of biased and unbiased networks to be compared. It is worth highlighting that the answer of biased agents is not included in the majority vote, and so the biased agents only affect the performance via the diffusion of their answers in the debate. In addition to analysing the effect of network structure and bias on the QA performance, the dynamics of the agent's responses and the level of consensus in the collective are also analysed. The code used to implement this work, as well as the agent's responses, are made publicly available: \href{https://www.github.com/tsukuba-websci/PSiLMN}{https://www.github.com/tsukuba-websci/PSiLMN}.

\section{Results}
\subsection{Network Structure and QA Performance}

Comparing the QA performance between different types of networks, it is evident that structure plays a role in the accuracy, as shown in Table~\ref{tab:unbiased_accuracy_comparison}. In particular, random networks achieve similar performance to fully connected networks, while using $250$ times fewer input tokens per round of debate. Scale-free networks, which use a similar amount of input tokens per round to random networks, exhibit worse performance than random networks, suggesting the random network topology is superior for problem-solving tasks. In contrast, fully disconnected networks demonstrate the lowest performance, highlighting the importance of collaborative problem-solving.

\begin{table}[htbp]
    \centering
    \begin{tabular}{lll}
        \toprule
        \textbf{Network}    & \textbf{Tokens per Round} & \textbf{Accuracy} \\
        \midrule
        Fully Connected     & 125000 & $ 67.7 \pm 1.1 \% $ \\
        Fully Disconnected  & 5000   & $ 63.9 \pm 0.4 \% $ \\
        Random              & 28600  & $ 68.2 \pm 0.5 \% $ \\
        Scale-Free          & 21800  & $ 64.8 \pm 1.0 \% $ \\
        \bottomrule
    \end{tabular}
    \caption{The accuracy and number of tokens used per round of debate for various types of (unbiased) networks.}
    \label{tab:unbiased_accuracy_comparison}
\end{table}

\begin{figure}[]
    \centering
    \includegraphics[width=0.9\linewidth]{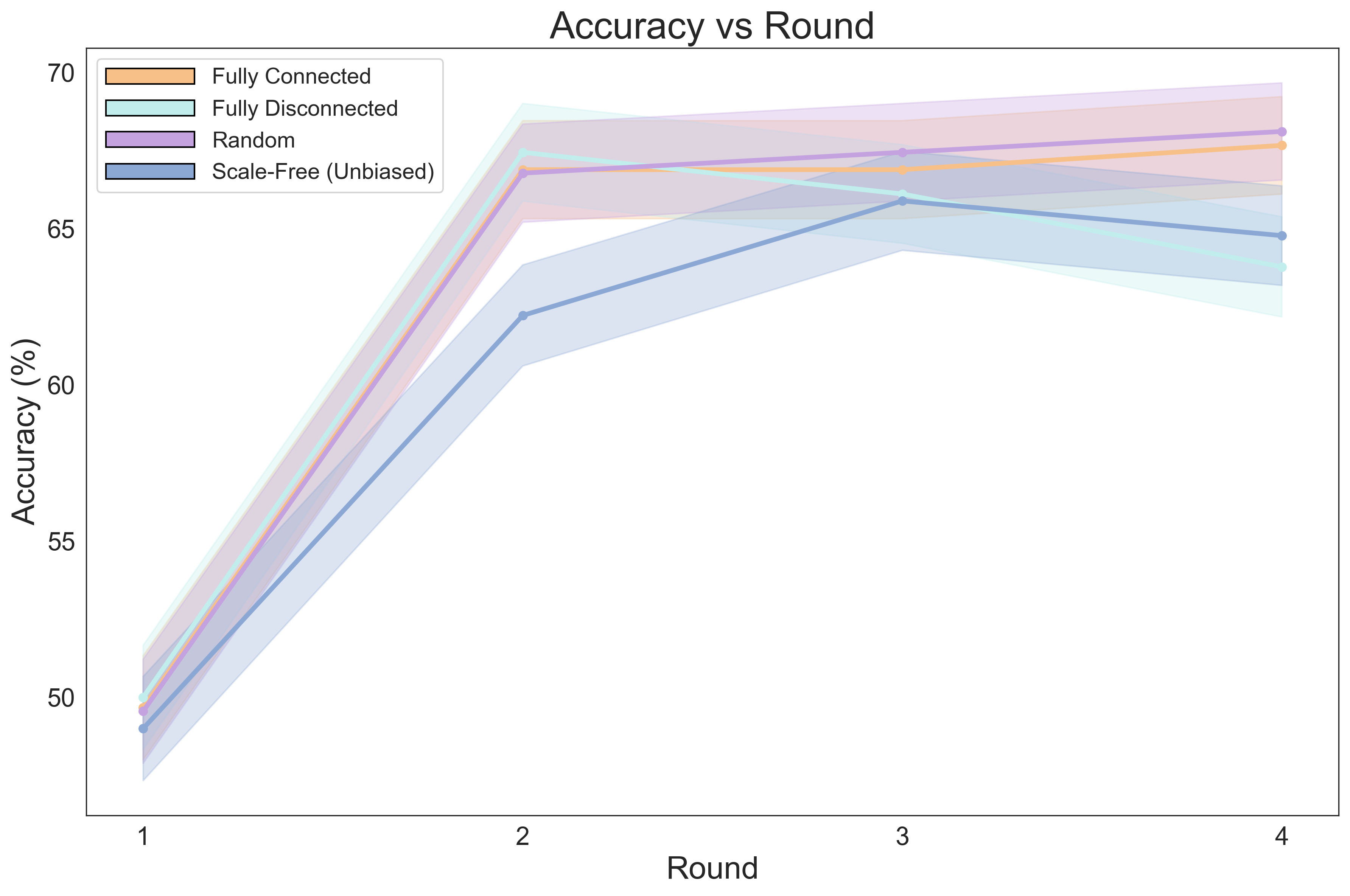}
    \caption{Accuracy per round of debate for different types of networks.}
    \label{fig:accuracy_vs_round_structure}
\end{figure}

To understand how the QA performance of the system evolves with each round of debate, a majority vote was taken after each round and the accuracy was computed, shown in Fig~\ref{fig:accuracy_vs_round_structure}. Initially, the accuracy of all systems is 50\% as the agents are yet to interact and influence each other. This performance can be regarded as a single-agent baseline. In the next round, the accuracy for all network types rises, with the steepest increase for fully connected networks, as the correct answers become distributed around the network. Notably, the accuracy of fully disconnected networks begins to decrease after round 2, which can be qualitatively attributed to the responses decreasing in quality throughout the debate. Inspecting the output of agents in these networks reveals short responses such as ``(B) 64'' or ``(D) 60'', with no reasoning provided in the final rounds of debate.

\subsection{Bias and QA Performance}

Comparing the performance between biased and unbiased systems, it is found that bias also plays a role in QA accuracy, shown in Table~\ref{tab:biased_accuracy_comparison}.

\begin{table}[htbp]
    \centering
    \begin{tabular}{llll}
        \toprule
        \textbf{Network}    & \textbf{Accuracy} \\
        \midrule
        Unbiased                     & $ 64.8 \pm 1.0 \% $ \\
        Correctly Biased (Hub)       &  $ 88.1  \pm 0.5 \% $ \\
        Incorrectly Biased (Hub)     & $ 43.8 \pm 1.5 \% $ \\
        Correctly Biased (Edge)      & $ 65.7 \pm 1.1 \% $ \\
        Incorrectly Biased (Edge)    & $ 64.9 \pm 1.3 \% $\\
        \bottomrule
    \end{tabular}
    \caption{Accuracy for biased scale-free networks.}
    \label{tab:biased_accuracy_comparison}
\end{table}

Networks with correctly (incorrectly) biased nodes at their hubs perform significantly better (worse) than their unbiased counterpart. In particular, networks with correctly biased hub nodes performed twice as well when compared to networks with incorrectly biased hubs, with accuracies of $88.1 \pm 0.5 \%$ and $43.8 \pm 1.5 \%$ respectively. Although bias is expected to impact the performance, the significant decrease in accuracy for incorrectly biased networks highlights that it only takes a few biased and well-connected agents, two in this case, to impair the results significantly. Moreover, the stronger comparative performance of the unbiased system demonstrates that although agents may be capable of solving problems correctly, they are easily influenced by incorrect agents. In the case where bias is inserted on the edge of the network on the other hand, it is found that there is little effect on the QA performance.

\begin{figure}[t!]
    \centering
    \includegraphics[width=0.9\linewidth]{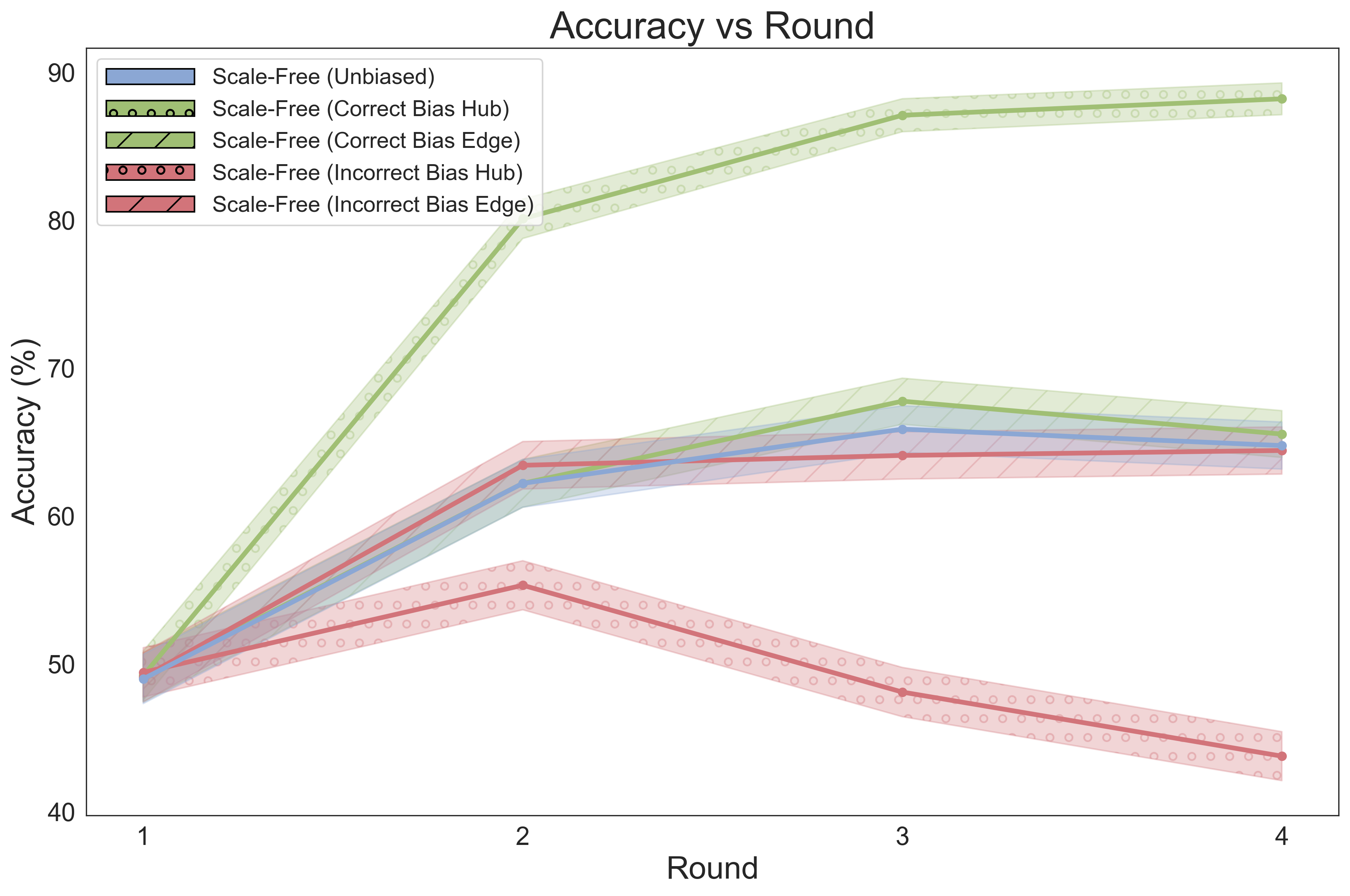}
    \caption{Accuracy per round of debate for scale-free networks with different types of biases.}
    \label{fig:accuracy_vs_round_bias}
\end{figure}

The accuracy for each round of debate in biased systems was found, shown in Fig~\ref{fig:accuracy_vs_round_bias}. As in Fig~\ref{fig:accuracy_vs_round_structure}, the accuracy for all systems is the same in round one and increases in round two, with the largest increase for the system with correctly biased hubs. This increase from $50 \%$ to $80\%$ accuracy can be attributed to the hub nodes communicating the correct answer to their neighbours, which comprise a large proportion of the network. Although the incorrectly biased hub nodes also communicate the incorrect answer to a large proportion of the network, the system's accuracy increases in round two. In subsequent rounds, however, the accuracy of the incorrectly biased hub networks decreases, which can be attributed to incorrect answers spreading around the network.

\subsection{Influence}

\begin{figure}[]
    \centering
    \includegraphics[width=0.95\linewidth]{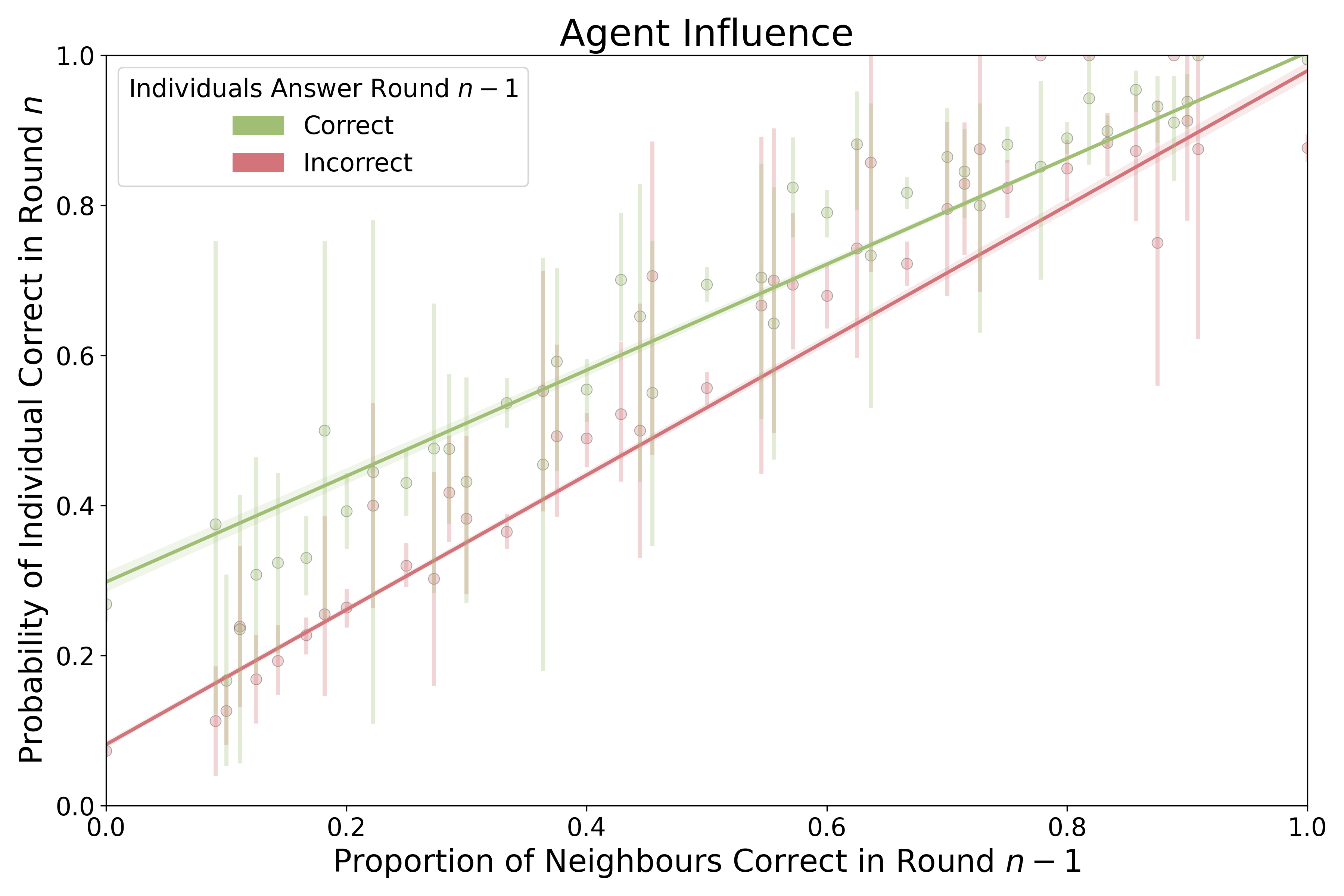}
    \caption{Agent influence; the likelihood of an agent being correct based on both its own answer and the answer of its neighbours in the previous round. The x-axis denotes the proportion of neighbours correct in the previous round, and the y-axis denotes the probability of the agent being correct in the current round. Points in green (red) indicate that the agent was correct (incorrect) in the previous round.}
    \label{fig:neighbour-accuracy}
\end{figure}

\begin{figure}[t!]
    \centering
    \begin{tabular}{cc}
        \begin{subfigure}[b]{0.45\columnwidth}
            \centering
            \includegraphics[width=\textwidth]{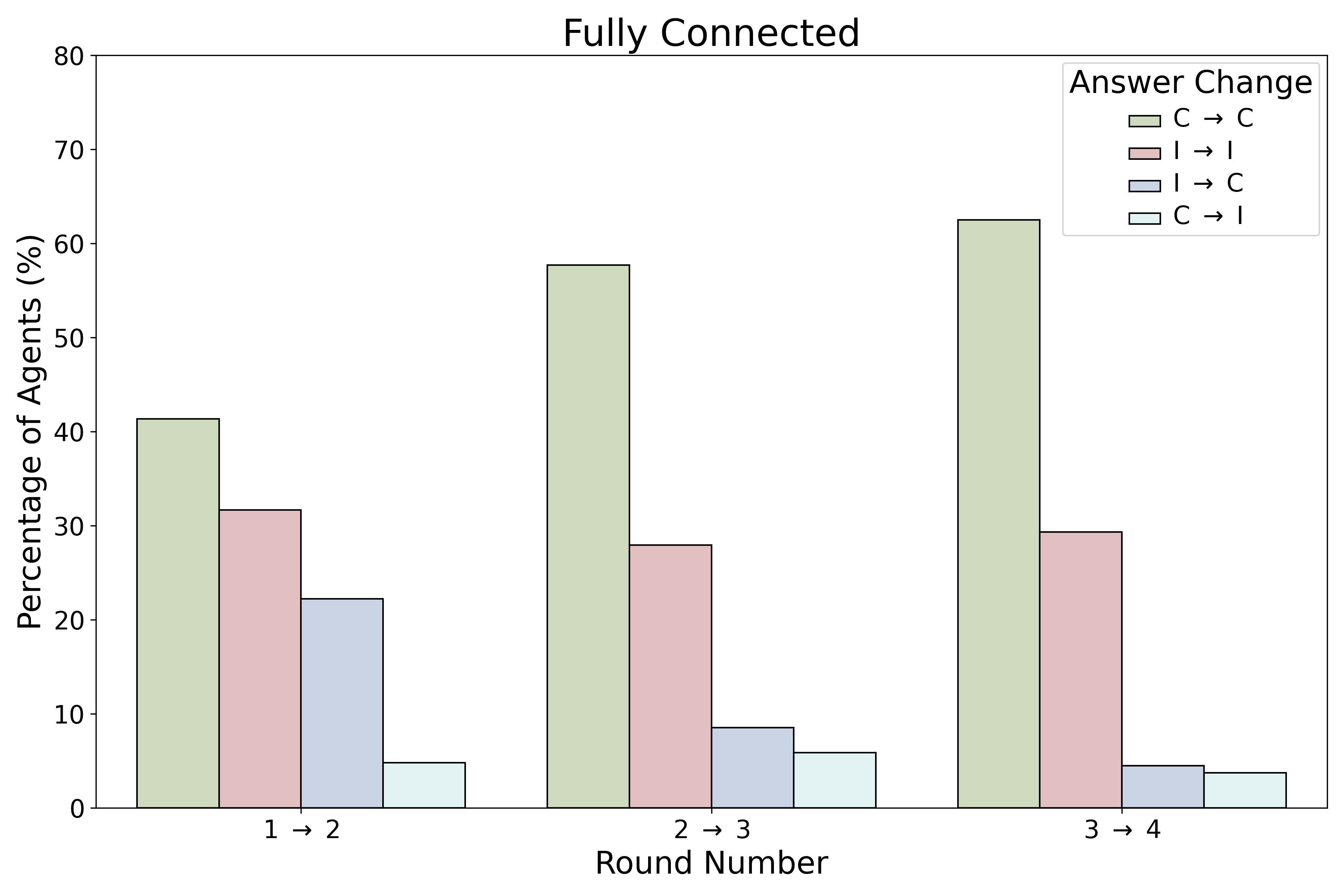}
            \caption{Fully Connected}
            \label{fig:overall-fully-connected}
        \end{subfigure} &
        \begin{subfigure}[b]{0.45\columnwidth}
            \centering
            \includegraphics[width=\textwidth]{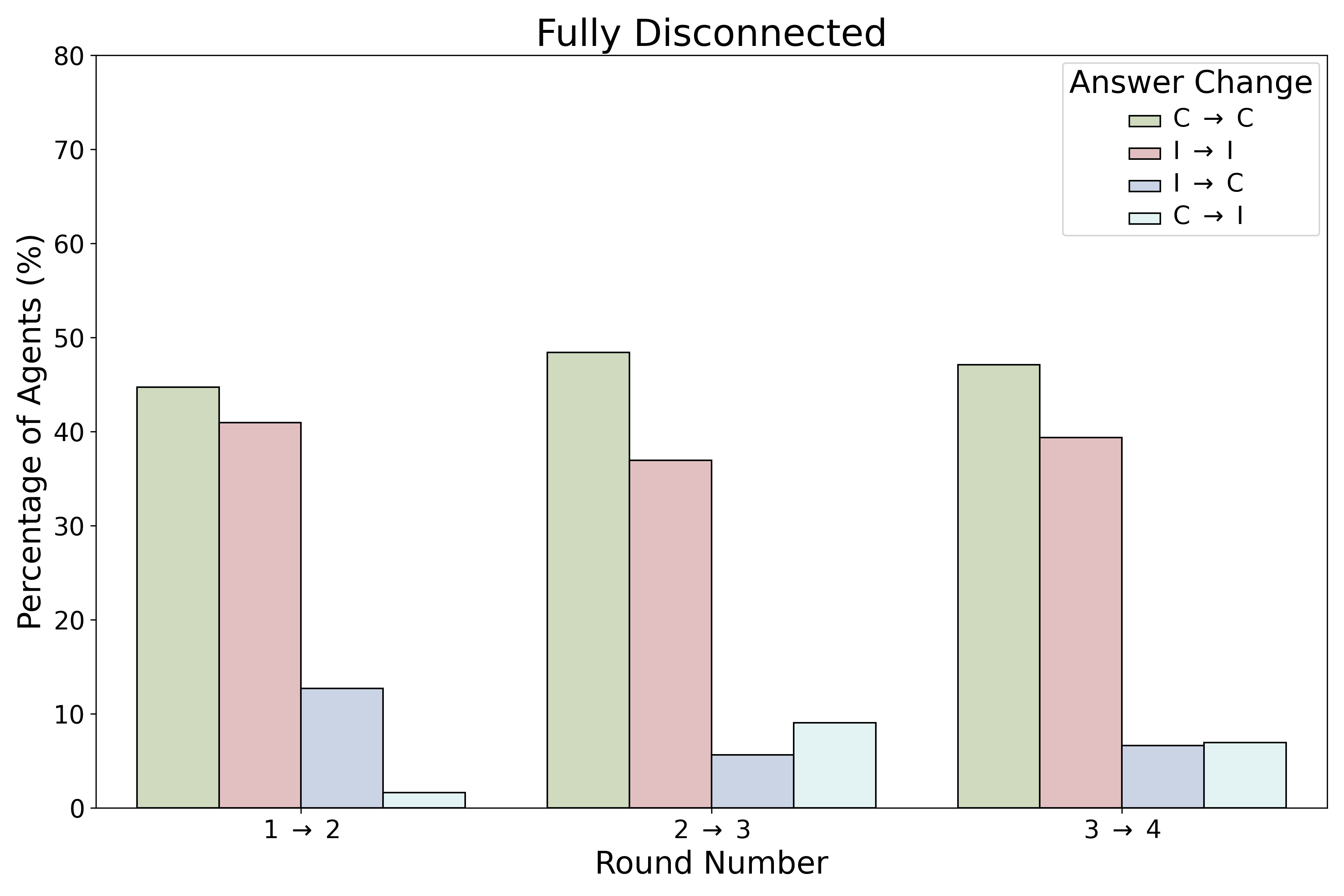}
            \caption{Fully Disconnected}
            \label{fig:overall-fully-disconnected}
        \end{subfigure} \\
        
        \begin{subfigure}[b]{0.45\columnwidth}
            \centering
            \includegraphics[width=\textwidth]{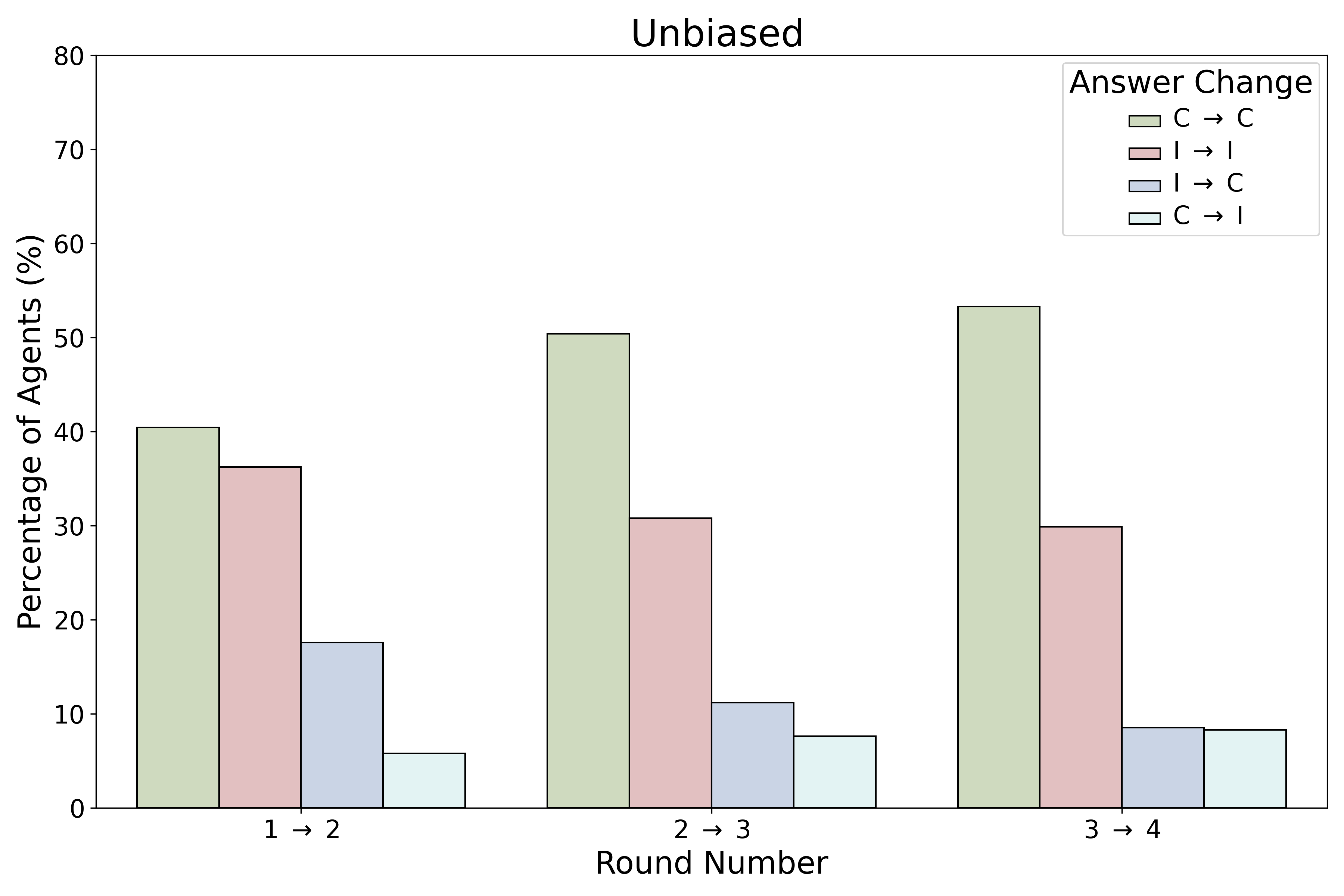}
            \caption{Scale-Free (Unbiased)}
            \label{fig:overall-unbiased}
        \end{subfigure} &
        \begin{subfigure}[b]{0.45\columnwidth}
            \centering
            \includegraphics[width=\textwidth]{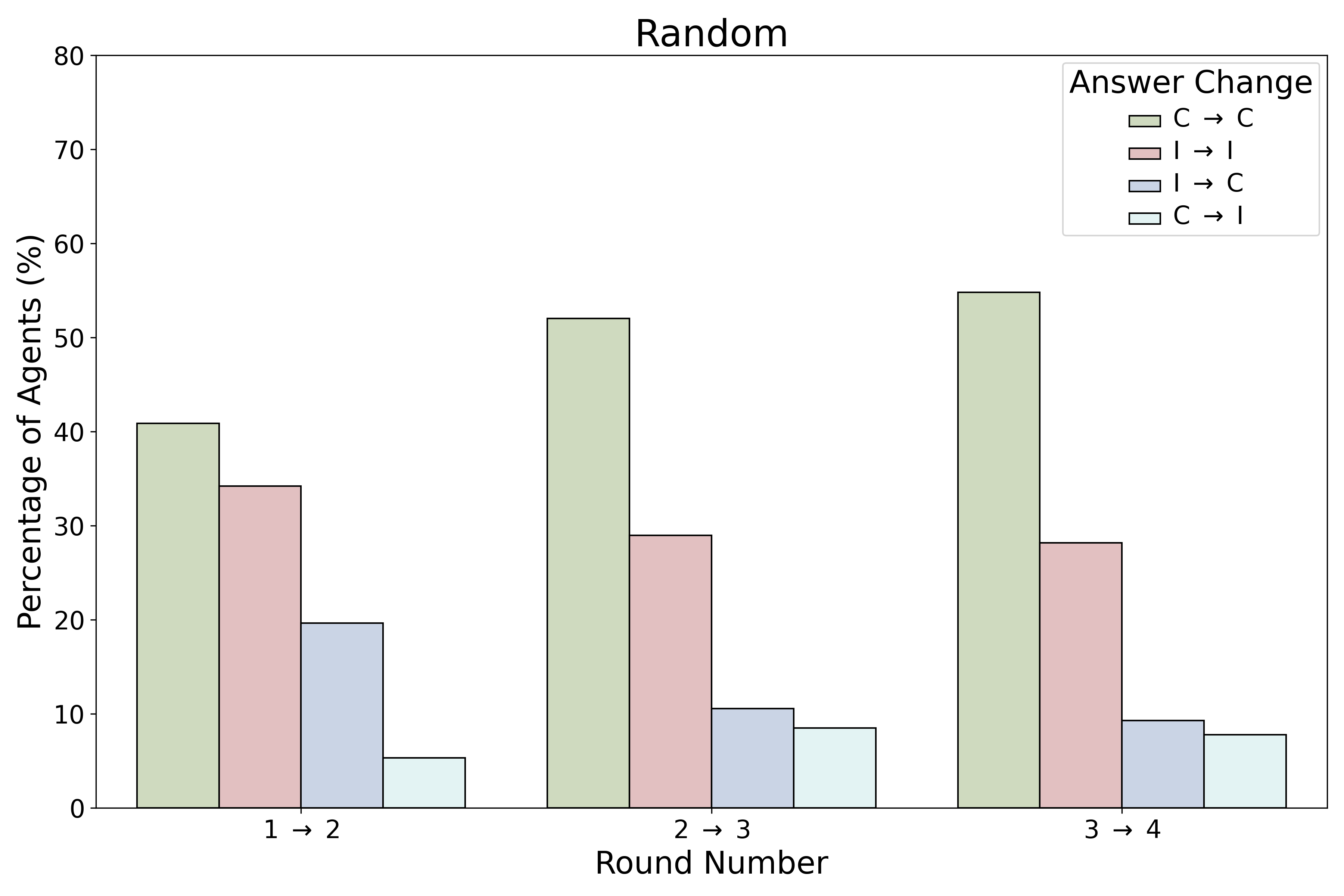}
            \caption{Random}
            \label{fig:overall-random}
        \end{subfigure} \\
        
        \begin{subfigure}[b]{0.45\columnwidth}
            \centering
            \includegraphics[width=\textwidth]{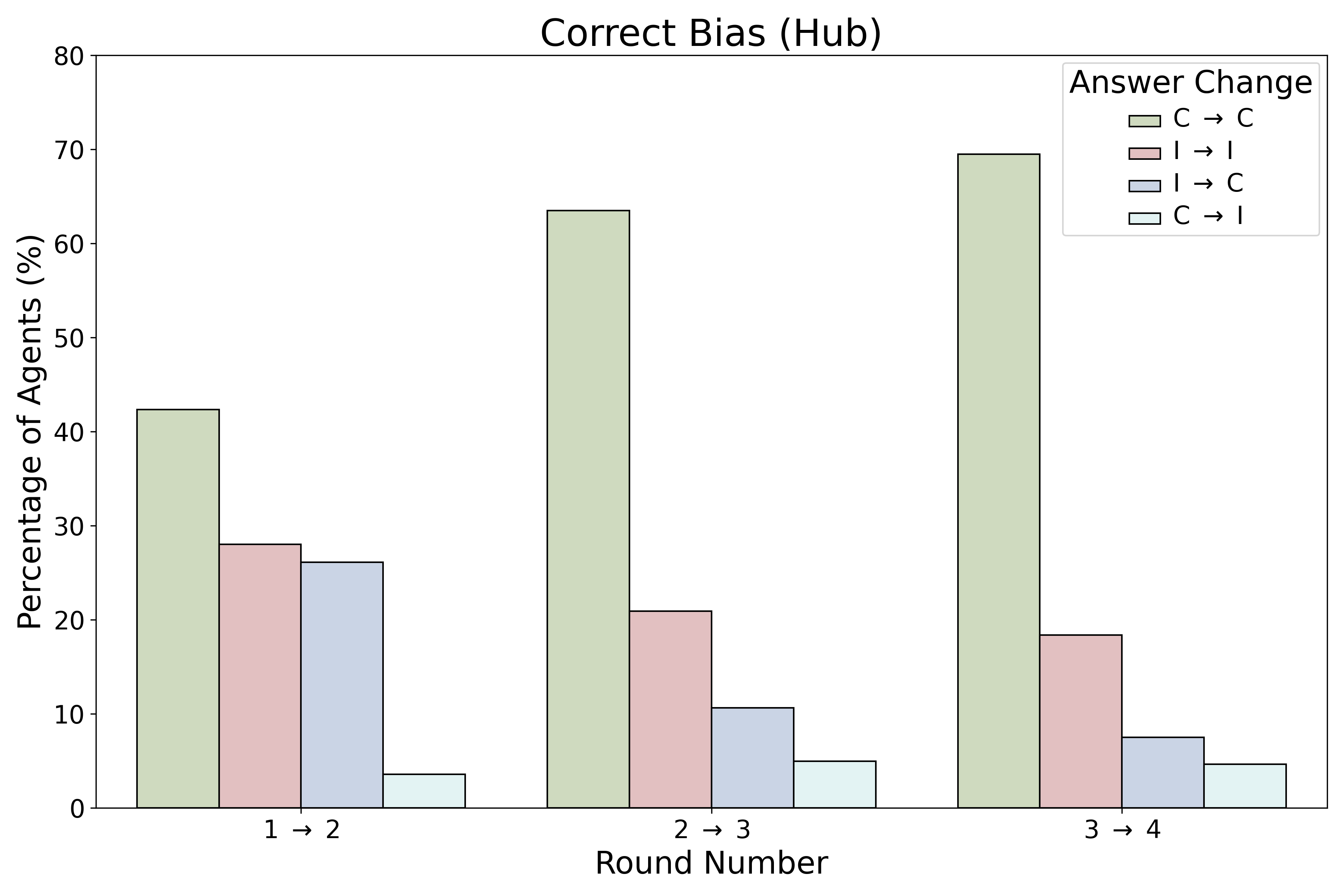}
            \caption{Correctly Biased (Hub)}
            \label{fig:overall-correct-bias-hub}
        \end{subfigure} &
        \begin{subfigure}[b]{0.45\columnwidth}
            \centering
            \includegraphics[width=\textwidth]{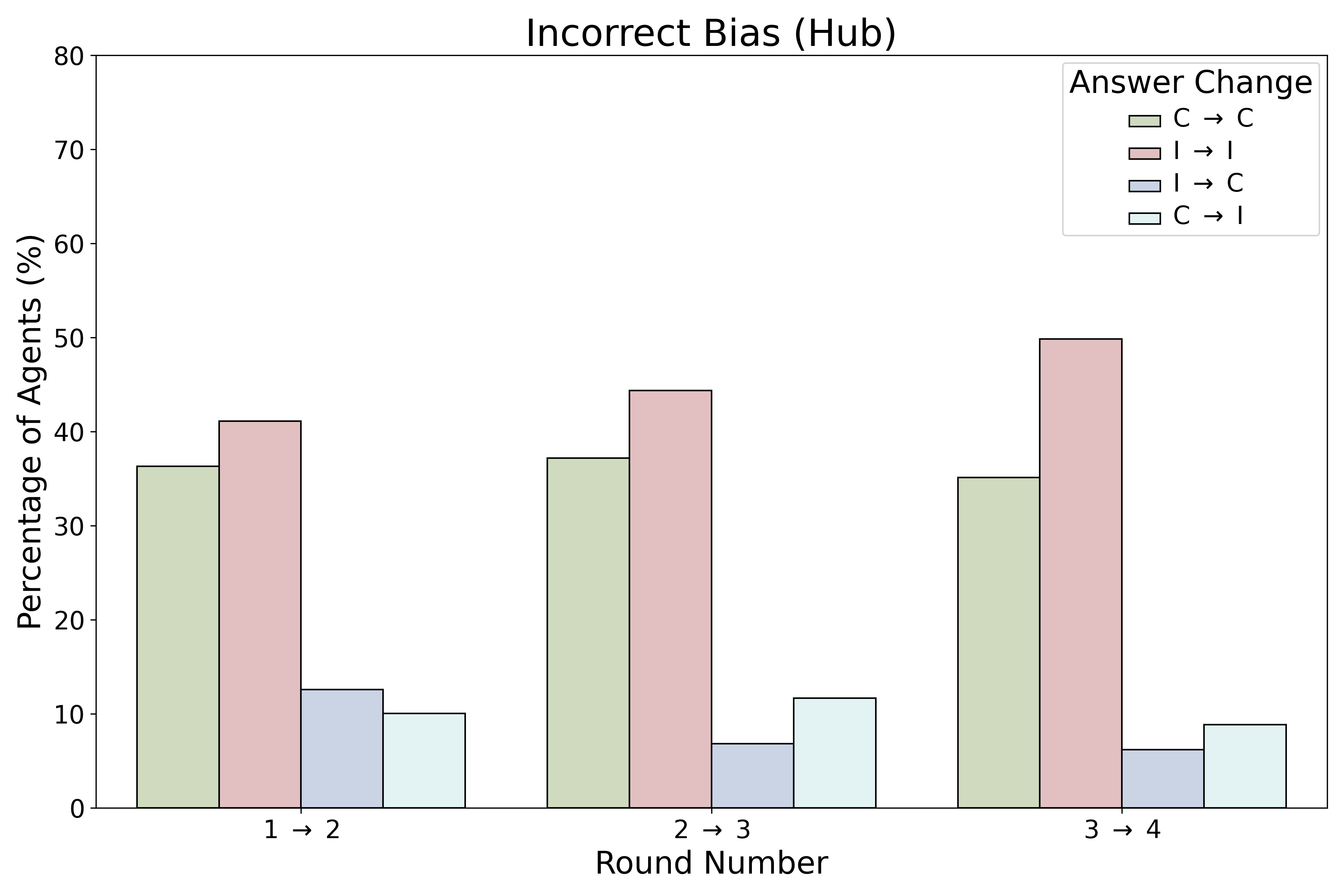}
            \caption{Incorrectly Biased (Hub)}
            \label{fig:overall-incorrect-bias-hub}
        \end{subfigure} \\
        
        \begin{subfigure}[b]{0.45\columnwidth}
            \centering
            \includegraphics[width=\textwidth]{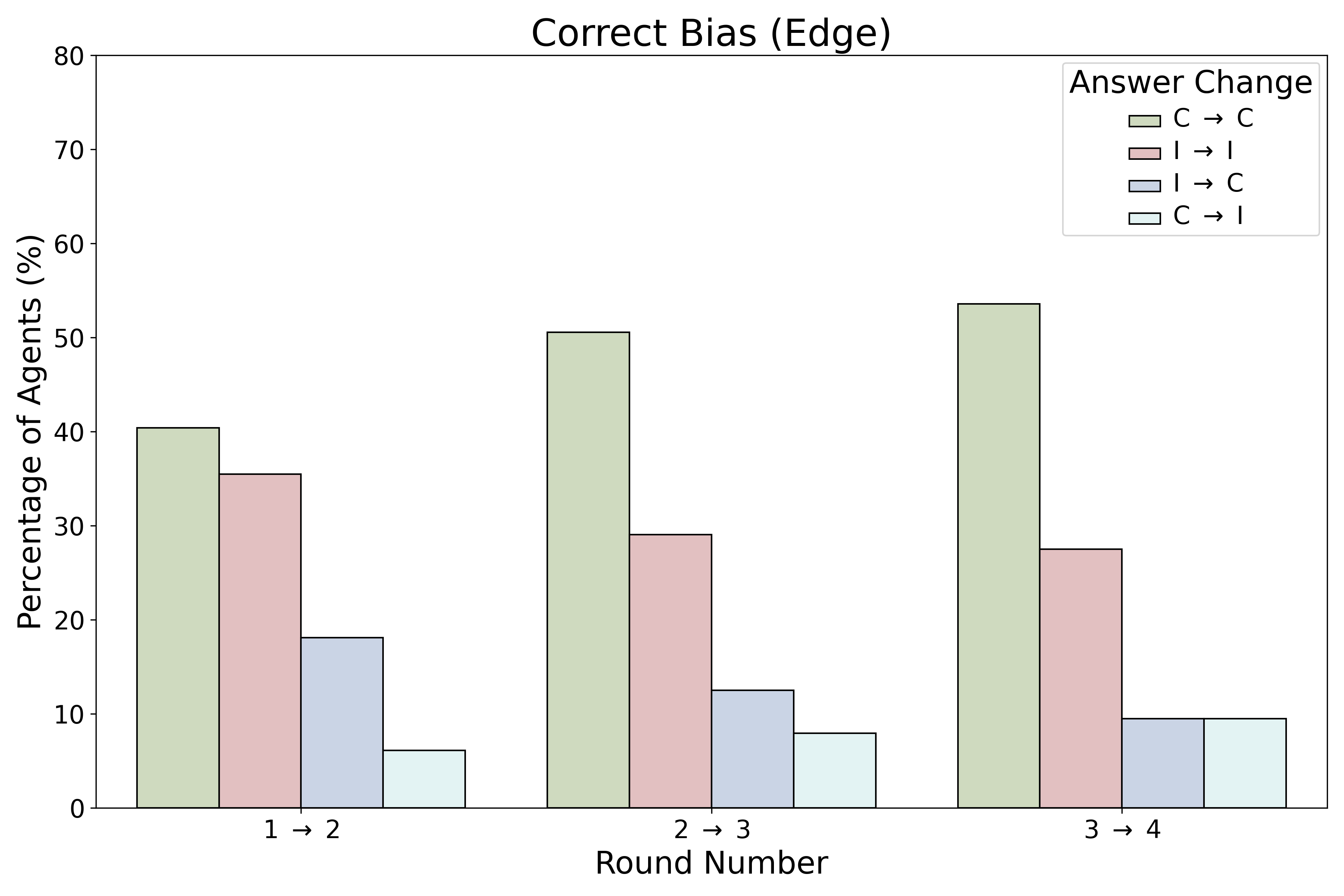}
            \caption{Correctly Biased (Edge)}
            \label{fig:overall-correct-bias-edge}
        \end{subfigure} &
        \begin{subfigure}[b]{0.45\columnwidth}
            \centering
            \includegraphics[width=\textwidth]{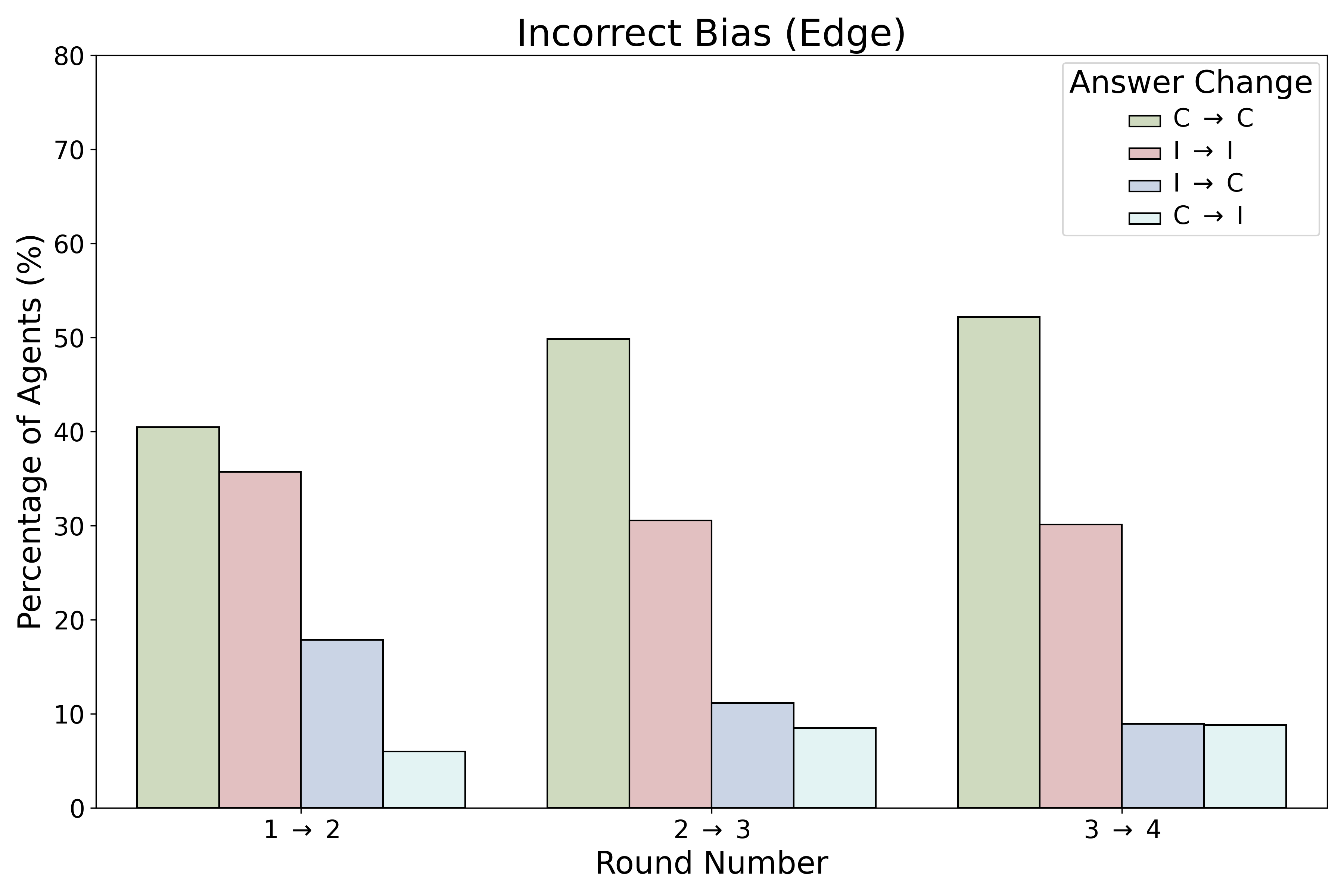}
            \caption{Incorrectly Biased (Edge)}
            \label{fig:overall-incorrect-bias-edge}
        \end{subfigure}
    \end{tabular}
    \caption{Answer changes for different networks and biases are illustrated. Each subplot shows the changes in answers over each round of debate. The number of agents remaining correct is indicated in green, while those remaining incorrect are indicated in red. Dark blue represents the number of agents switching from incorrect to correct, and light blue represents those switching from correct to incorrect.}
    \label{fig:opinion-changes}
\end{figure}

To understand how an agent may be influenced, the probability of the agent being correct in round $n$, given its previous response and the response of its neighbours in round $n-1$ is shown in Fig~\ref{fig:neighbour-accuracy}, with green (red) points showing cases where the agents own response was correct (incorrect) in the previous round. For clarity, the figure depicts the case of random networks, where each node will have, on average, the same number of neighbours. The figure shows that as the number of correct neighbours increases, so too does the probability of the agent being correct. Furthermore, the tendency for green points to lie above red points highlights the positive impact of self-reflection; regardless of the neighbours' responses, an agent is more likely to answer correctly if it was correct in the previous round.

Considering the edge cases reveals several interesting insights about the agent's behaviour. Firstly, when the agent and all of its neighbours were incorrect in the previous round, there is a $10\%$ chance that the agent will spontaneously switch to the correct answer. If the agent was correct on the other hand, this probability increases to $30\%$ demonstrating the positive effects of self-reflection. Secondly, when an agent was incorrect in the previous round, but is surrounded by correct agents, it is very unlikely to remain incorrect, showing a decrease in the impact of self-reflection. Finally, in the case where the neighbours' responses are split, the agent is equally likely to be correct or incorrect, with a slight increase for agents correct in the previous round. These findings highlight the importance of both individuality and collective thinking in multi-agent systems. That is, collaborative problem-solving improves the overall performance of the collective, while self-reflection acts to improve performance when local interactions are misguided.

\subsection{Dynamics}

To further understand the dynamics of these systems,  the way in which agents change their answers between rounds is shown in Fig~\ref{fig:opinion-changes}. In the case of fully connected (Fig~\ref{fig:overall-fully-connected}), scale-free (Fig~\ref{fig:overall-unbiased}) and random (Fig~\ref{fig:overall-random}) networks, the number of agents selecting and remaining on the correct answer increases with each round of debate. For fully disconnected networks, on the other hand, the number of agents remaining correct or incorrect is near-constant, with agents continuing to switch between the correct and incorrect answers. This behaviour is unique to fully disconnected networks, illustrated by the light and dark blue bars in Fig~\ref{fig:overall-fully-disconnected}.

Considering bias, networks correctly biased at their hubs (Fig~\ref{fig:overall-correct-bias-hub}) exhibit a large number of agents switching from incorrect to correct answers after the first round, in agreement with Fig~\ref{fig:accuracy_vs_round_bias}. These agents with correct answers tend to keep the correct response throughout the remaining rounds of debate. Networks incorrectly biased at their hubs (Fig~\ref{fig:overall-incorrect-bias-hub}), on the other hand, have an increasing number of agents switching from correct to incorrect after round two. This is a significant result, as it highlights the fact that agents may have the correct answer, but will be convinced to switch due to the influence of their biased neighbours. Moreover, this suggests that multi-agent influence plays a larger role in performance than self-reflection, in general agreement with Fig~\ref{fig:neighbour-accuracy}. If self-reflection had a significant impact on the system then correct agents would disregard the responses of their incorrect neighbours, highlighting the importance of collaboration in the problem-solving domain. As expected, networks biased at their edges (Fig~\ref{fig:overall-correct-bias-edge} and Fig~\ref{fig:overall-incorrect-bias-edge}) have similar dynamics to the unbiased counterpart.

\subsection{Consensus}
While the accuracy gave us an insight into the average QA performance of the system, it provides little information on how the answers are distributed inside the network during any given round and whether or not the agents agree. In fact, the network can be \textit{correct} with less than half of its agents giving the correct answer, due to majority voting. This section explores how and under which conditions a consensus is formed.

\begin{figure}[]
    \centering
    \includegraphics[width=0.9\linewidth]{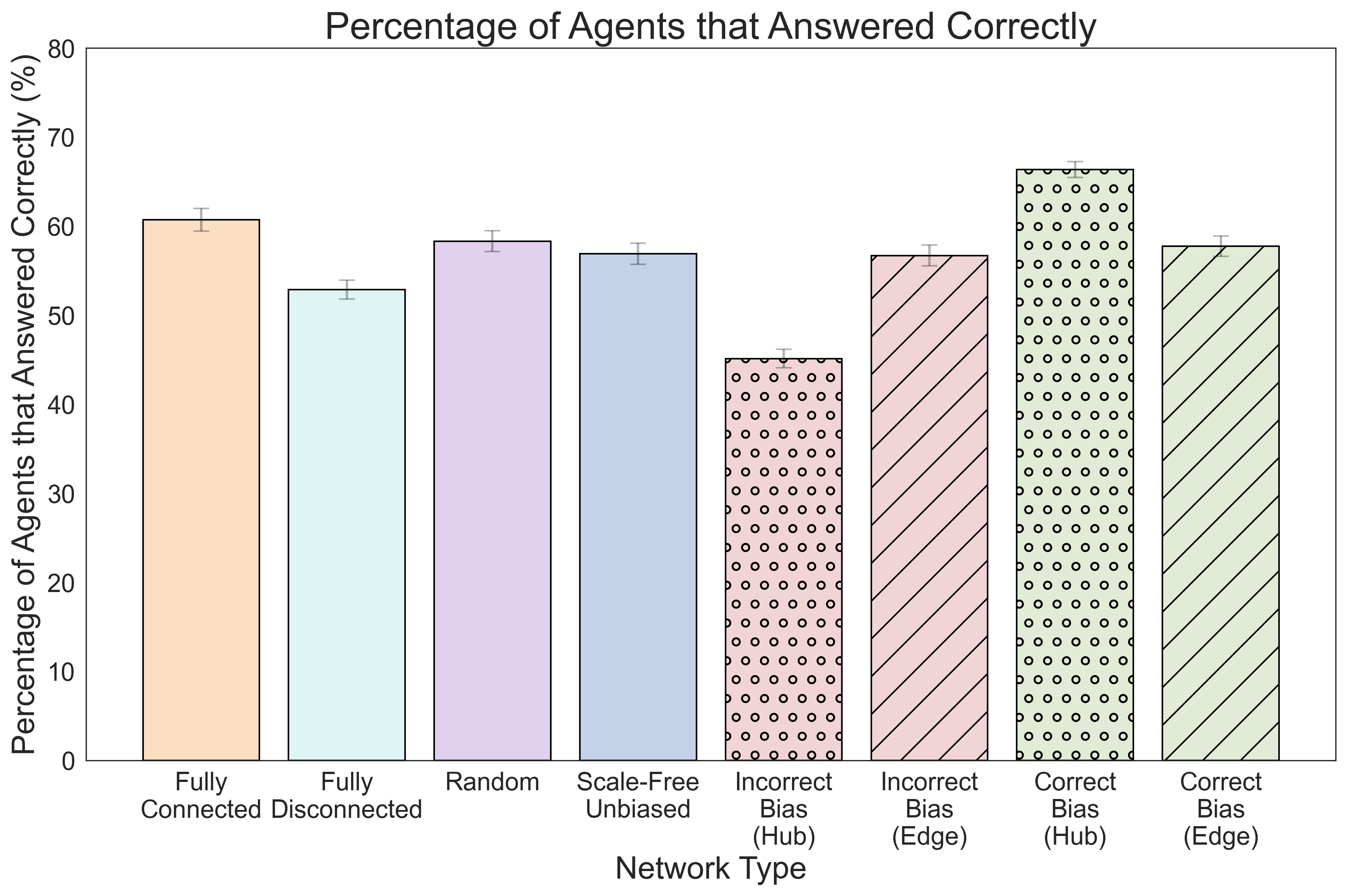}
    \caption{The percentage of agents that answered correctly in the final round.}
    \label{fig:correct_prop_vs_bias}
\end{figure}

Fig~\ref{fig:correct_prop_vs_bias} shows the percentage of agents in the network that answered the question correctly in the final round. Although this metric highlights the relationship between the consensus towards the correct answer and the overall QA performance, which captures the total number of questions answered correctly (Table ~\ref{tab:unbiased_accuracy_comparison} and ~\ref{tab:biased_accuracy_comparison}), little information is provided on how the answers are distributed. 

To gain insight into the distribution of answers, the Simpson index is used to estimate the level of consensus within the collective~\citep{simpson1949measurement}.  The Simpson index $\lambda$, which is used to quantify diversity, measures the probability that any two randomly selected agents give the same answer in the final round of the experiment and is given by Simpson's formula:

$$\lambda = \sum_{i \in R}^{}p_{i}^2$$

\noindent where $R$ is the number of possible responses and $p_i$ is the proportional abundance of answer type $i$. 

In this experiment, agents can give five different types of responses: A, B, C, D and ``Undetermined''. A, B, C and D naturally correspond to the different multiple-choice options of the MMLU dataset and ``Undetermined'' refers to the situation where the agent fails to provide a valid option. In this approach, the minimum Simpson index, which occurs when answers are evenly split between all five types of responses, is given by $\lambda_{\text{min}} = 5 \times 0.2^2 = 0.2$. On the other hand, if all agents agree on the same answer, the Simpson index is given by $\lambda_{\text{max}} = 1$, regardless of whether the answer is correct or incorrect. Moreover, if the responses are completely split, for example, if half the agents answer A and the other half answer B, then the Simpson index is $\lambda_\text{split} = 0.5$. This consensus measure allows us to remove the separation between correct and incorrect answers, and to measure exclusively the consensus between agents.

The average Simpson index for each network type is shown in the ``overall'' column of Table~\ref{tab:network_simpson}. High values for fully connected networks, followed by those for random and scale-free networks, indicate a relationship between network connectivity and the agreement among agents.

\begin{table}[htbp]
    \centering
    \begin{tabular}{llll}
        \toprule
        \textbf{Network} & \textbf{Overall} & \textbf{Correct} & \textbf{Incorrect} \\
        \midrule
        Fully Connected           & $0.710$ & $0.775$ & $0.574$ \\
        Fully Disconnected        & $0.529$ & $0.618$ & $0.371$ \\
        Random                    & $0.628$ & $0.708$ & $0.457$ \\
        Scale-Free \\
        \ \ Unbiased    & $0.623$ & $0.712$ & $0.459$ \\
        \ \ Correct (Hub)    & $0.613$ & $0.641$ & $0.405$ \\
        \ \ Incorrect (Hub)  & $0.506$ & $0.633$ & $0.407$ \\
        \ \ Correct (Edge)   & $0.613$ & $0.703$ & $0.439$ \\
        \ \ Incorrect (Edge) & $0.614$ & $0.700$ & $0.455$ \\
        \bottomrule
    \end{tabular}
    \caption{Overall, correct, and incorrect Simpson consensus values for various network and bias types.}
    \label{tab:network_simpson}
\end{table}

The following question then arises; ``when the system answers correctly, do they tend to agree on the same correct answer, or are the responses split?''. To answer this, the Simpson index $\lambda$ was calculated separately for questions answered correctly and incorrectly by the system and is depicted in the ``correct'' and ``incorrect'' columns of  Table~\ref{tab:network_simpson}. In particular, we see that when the system answers correctly, there is a higher consensus among agents, indicating greater certainty. Conversely, when the system answers incorrectly, the consensus is weaker, reflecting less certainty. Therefore, the degree of consensus among agents can be used to quantify the confidence in the correctness of an answer.

\begin{figure}[t!]
    \centering
    \begin{tabular}{cc}
        \begin{subfigure}[b]{0.22\textwidth}
            \centering
            \includegraphics[width=\textwidth]{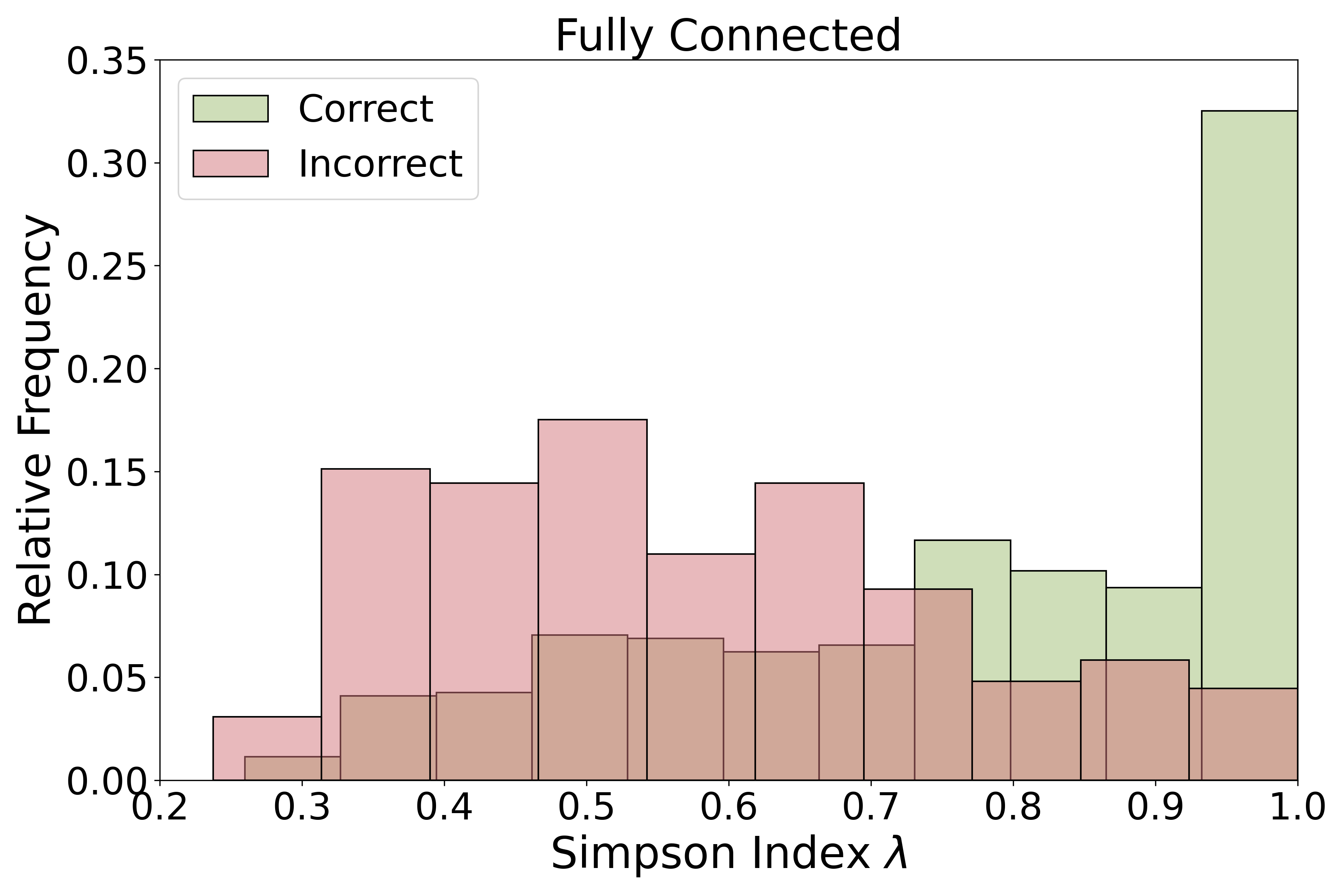}
            \caption{Fully Connected}
            \label{fig:simpson_fully_connected}
        \end{subfigure} &
        \begin{subfigure}[b]{0.22\textwidth}
            \centering
            \includegraphics[width=\textwidth]{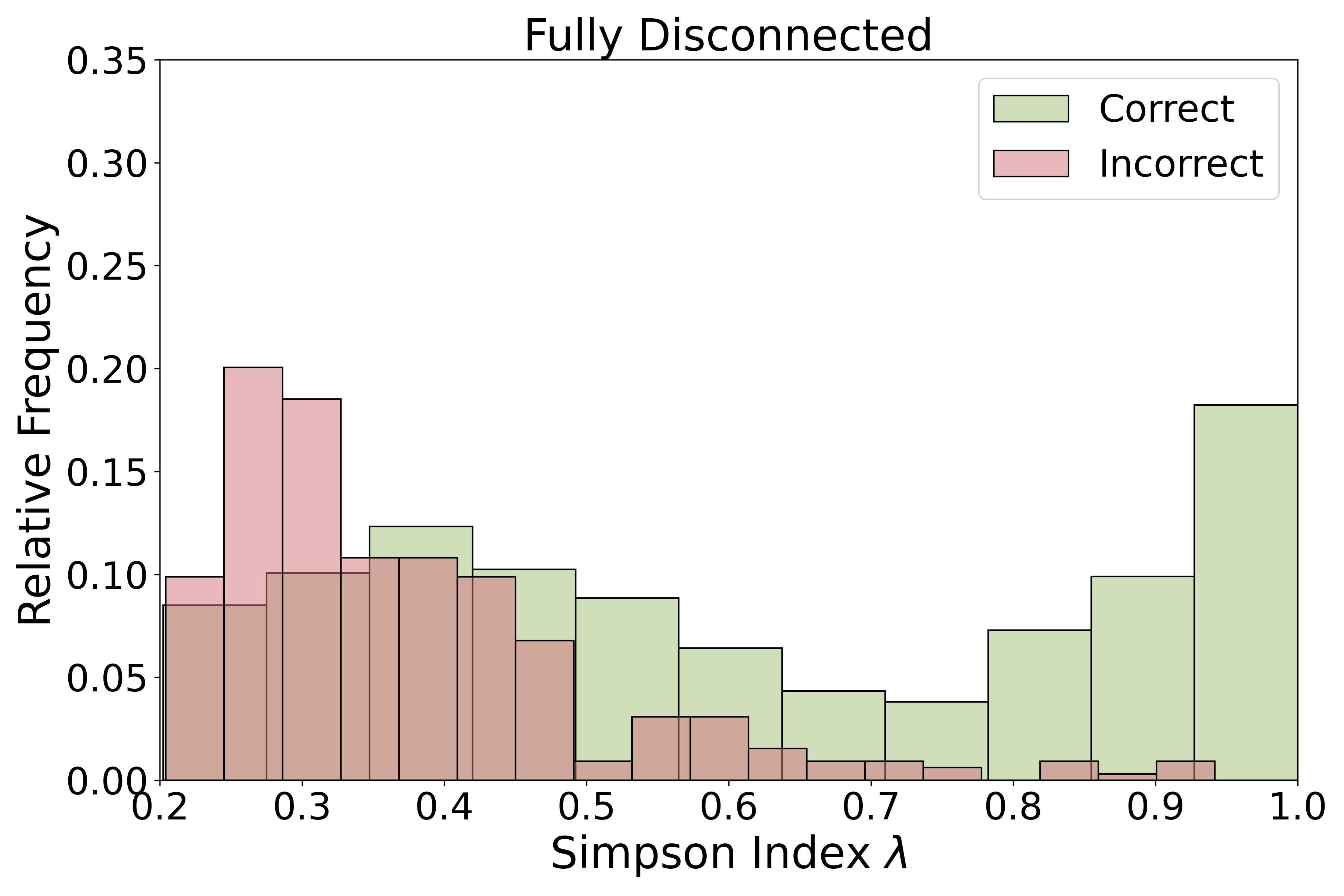}
            \caption{Fully Disconnected}
            \label{fig:simpson_fully_disconnected}
        \end{subfigure} \\
        
        \begin{subfigure}[b]{0.22\textwidth}
            \centering
            \includegraphics[width=\textwidth]{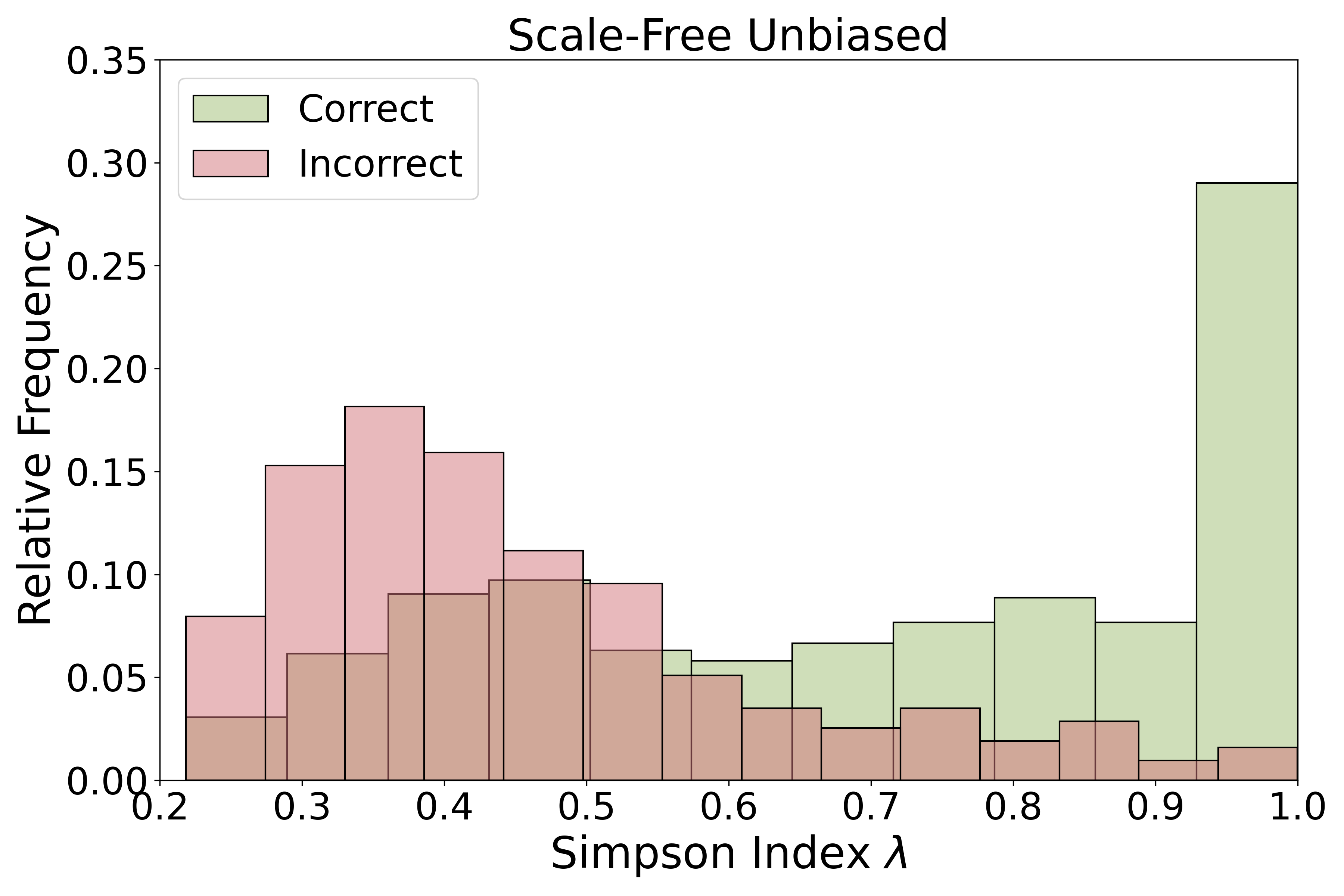}
            \caption{Scale-Free (Unbiased)}
            \label{fig:simpson_unbiased}
        \end{subfigure} &
        \begin{subfigure}[b]{0.22\textwidth}
            \centering
            \includegraphics[width=\textwidth]{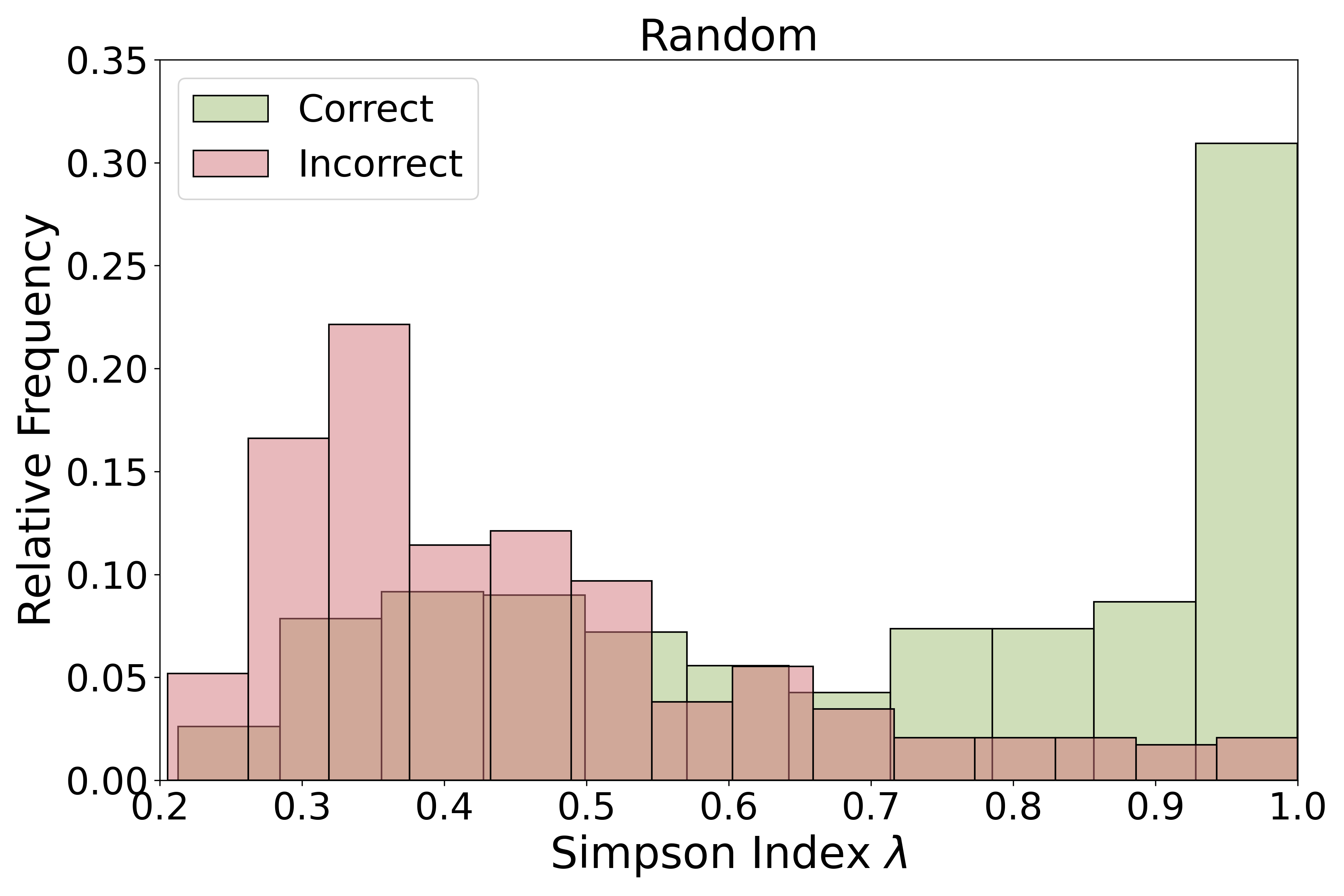}
            \caption{Random}
            \label{fig:simpson_random}
        \end{subfigure} \\
        
        
    \end{tabular}
    \caption{The distribution of the Simpson index for each question administered to the system, for different networks. Green (red) indicates questions the system answered correctly (incorrectly). The distribution of correct answers (green) tends to be left-skewed, highlighting a tendency for systems to form a consensus when the answer is correct. Conversely, the distribution is right-skewed for incorrect answers.}
    \label{fig:simpson_dist}
\end{figure}

This is further highlighted by Fig~\ref{fig:simpson_dist}, which plots the distribution of the Simpson index for different types of networks. In all networks, a strong consensus usually corresponds to correct answers, shown by the large green spike and few red points at $\lambda \approx 1$. This indicates that when the system reaches full consensus, it is likely correct. Conversely, split answers tend to correlate with incorrect responses.

Bias in networks, regardless of being correct or incorrect bias, slightly reduces the consensus among agents, as depicted in the ``correct'' and ``incorrect'' columns of Table~\ref{tab:network_simpson}. This result suggests that when the system answers incorrectly, the presence of bias leads to an increase in uncertainty amongst the agents, thus reducing the average level of consensus.

Overall, these results show that analysing consensus in networks reveals important insights into the agent's collective behaviour and provides guidance for designing networks for problem-solving tasks. The analysis highlights that a high degree of consensus among agents correlates with correct answers, indicating greater certainty. Conversely, when consensus is lower, the system is more likely to be incorrect. These findings highlight the importance of designing connected networks to facilitate the spreading of correct answers, allowing agents to form a consensus.

\section{Conclusion}
This work extends multi-agent debate to more complex network topologies by representing each agent as a node, connected to their debate partners through communication channel edges, with self-loops indicating agent self-reflection. To understand the QA performance and dynamics of these systems, MMLU questions were administered to scale-free, random, fully connected and fully disconnected networks.  Additionally, the effect of bias was analysed by manually inserting correct or incorrect responses into the hubs and edges of scale-free networks. 

The results show that random networks perform similarly to fully connected networks while using significantly fewer tokens. Fully disconnected networks, on the other hand, perform the poorest, with answers degrading with each round of debate. Additionally, bias plays a strong role in the QA capabilities, with a few incorrectly biased agents severely weakening the overall performance. Analysing how agents are influenced revealed several interesting behaviours, such as the importance of self-reflection when surrounded by incorrect neighbours and a weakening impact of self-reflection when the proportion of correct neighbours increases, highlighting the importance of both individuality and collaboration in the problem-solving domain. Furthermore, the results show a relationship between network connectivity and consensus, with answers likely to be correct when the collective is in full agreement, quantifying the uncertainty in the system.

\section{Discussion and Limitations}\label{sec:limit}

The findings of this study have significant implications for the design of future multi-agent systems, a topic of interest in the field of artificial life. The results suggest that when designing collective systems of LLM agents for problem-solving tasks, random networks provide a performant and cost-effective approach. Moreover, measures of consensus, such as the Simpson index, can be utilised as a useful tool for gauging the uncertainty in these systems. 

The effect of biasing scale-free networks also provides some insight into how we design future systems. For instance, while it has been shown that combining different models in multi-agent debate can lead to an increase in performance~\citep{debate}, the results of this study suggest that placing larger, more powerful models at the central hubs and smaller models at the periphery may lead to an increase in performance without the high computation cost of using a system comprised of large models alone. An analysis of this behaviour is left to future research.

A significant limitation in this work is the limited number of agents, questions, rounds of debate and networks used. As the number of requests to the OpenAI API increases significantly with each of these parameters, certain experimental configurations become prohibitively expensive. In this work, the parameters were selected as a trade-off between statistical significance and generality. In future work, it would be interesting to study other network structures such as small-world networks, and networks with dynamic topologies, as well as increasing the number of agents, which may lead to an increase in performance ~\citep{sayama2015effects}. Additionally, while this work focused solely on QA tasks, future research should explore other aspects of intelligence. For instance, creativity may also benefit from multi-agent discussions, as suggested by ~\citep{lu2024llm}.

\footnotesize
\bibliographystyle{apalike}
\bibliography{main} 

\end{document}